\documentclass[10pt,twocolumn,letterpaper]{article}

\usepackage{iccv}
\usepackage{times}
\usepackage{epsfig}
\usepackage{graphicx}
\usepackage{amsmath}
\usepackage{amssymb}
\usepackage{tikz}
\usepackage{cite}	

\newcommand{\todo}[1]{\textcolor{red}{[TODO: #1]}}
\newcommand{\vitto}[1]{\textcolor{red}{[VF: #1]}}
\newcommand{\rahul}[1]{\textcolor{blue}{[RS- #1]}}
\newcommand{\luca}[1]{\textcolor{orange}{[LUCA- #1]}}
\newcommand{\ricco}[1]{\textcolor{purple}{[RICCO- #1]}}
\newcommand{\google}[1]{\textcolor{red}{[GOOGLE- #1]}}
\renewcommand{\todo}[1]{}
\renewcommand{\vitto}[1]{}
\renewcommand{\rahul}[1]{}
\renewcommand{\luca}[1]{}
\renewcommand{\ricco}[1]{}
\renewcommand{\google}[1]{}

\DeclareMathAlphabet{\mathpzc}{T1}{pzc}{m}{n}

\newlength{\halfwidth}
\newlength{\fullwidth}
\usetikzlibrary{shapes}
\pgfdeclarelayer{background}
\pgfsetlayers{background,main}
\newlength{\tikzimgheight}
\newlength{\tikzimgwidth}

\usepackage{ifthen}
\usetikzlibrary{calc}

\usepackage[pagebackref=true,breaklinks=true,letterpaper=true,colorlinks,bookmarks=false]{hyperref}

\iccvfinalcopy 

\def\iccvPaperID{1200} 

\ificcvfinal\fi
\begin{document}

\title{Recovering Spatiotemporal Correspondence between Deformable Objects by Exploiting Consistent Foreground Motion in Video}

\author{Luca Del Pero\textsuperscript{1} \hspace{1.22cm} Susanna Ricco\textsuperscript{2} \hspace{1.22cm} Rahul Sukthankar\textsuperscript{2} \hspace{1.22cm} Vittorio Ferrari\textsuperscript{1}\\
{\tt\small \hspace{-0.6cm} ldelper@inf.ed.ac.uk \hspace{0.5cm} ricco@google.com \hspace{0.7cm} sukthankar@google.com  \hspace{0.1cm} ferrari@inf.ed.ac.uk}\\
\hspace{-0.6cm} \textsuperscript{1}University of Edinburgh \hspace{0.8cm}\textsuperscript{2}Google Research
}
\maketitle

\begin{abstract}
  Given unstructured videos of deformable objects, we automatically recover spatiotemporal correspondences to map one object to another (such as animals in the wild). While traditional methods based on appearance fail in such challenging conditions, we exploit consistency in object motion between instances. Our approach discovers pairs of short video intervals where the object moves in a consistent manner and uses these candidates as seeds for spatial alignment. We model the spatial correspondence between the point trajectories on the object in one interval to those in the other using a time-varying Thin Plate Spline deformation model.
  On a large dataset of tiger and horse videos, our method automatically aligns thousands of pairs of frames to a high accuracy, and outperforms the popular SIFT Flow algorithm.
\end{abstract}

\section{Introduction}
\label{sec:introduction}

\begin{figure}
\begin{center}
\includegraphics[scale =0.42]{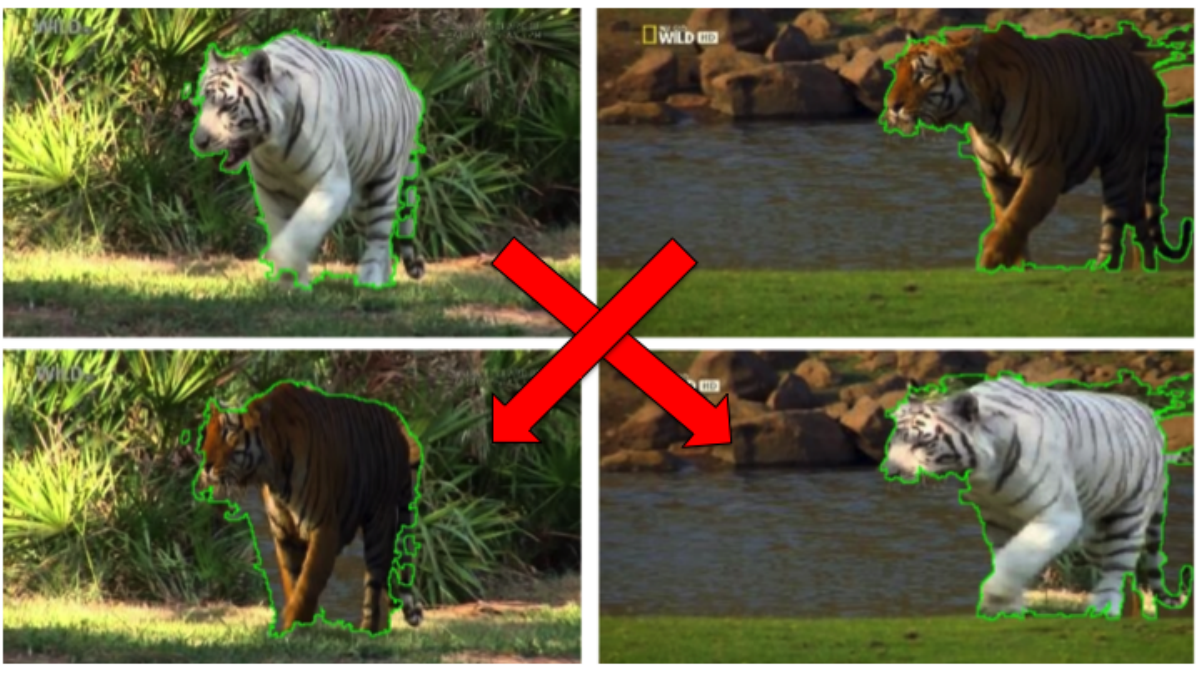}
\end{center}
 \caption{\small{
We recover point-to-point spatiotemporal correspondences across a collection
of unstructured videos of deformable objects.
Here, we display the recovered correspondences by mapping tigers from frames in two different
videos (top) onto each other (bottom).
Our method maps each part of a tiger in the first video to the corresponding part in the second
(\eg, head goes to head, front-right paw goes to front-right paw, etc.).
We use motion cues to find short video intervals where the foreground moves in a consistent manner.
This enables finding correspondences despite large variations in appearance (\eg, white and orange tigers).
}}
\vspace{-4pt}
\label{fig:teaser}
\end{figure}

\begin{figure*}[t]
\begin{center}
\setlength{\tabcolsep}{0.8pt}
\begin{tabular}{c c c c c c}
\footnotesize{\textbf{foreground masks}} & \footnotesize{\textbf{trajectory matches}} & \footnotesize{\textbf{homography}} & \footnotesize{\textbf{TPS mapping}} &  \footnotesize{\textbf{foreground edge points}} \\
\includegraphics[scale =0.25]{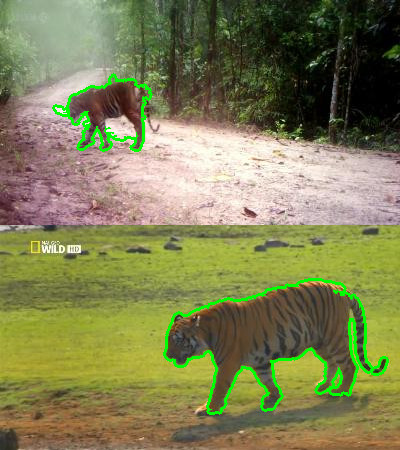}  &
\includegraphics[scale =0.25]{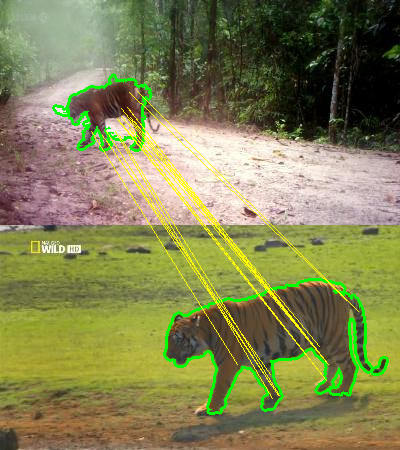}  &
\includegraphics[scale =0.25]{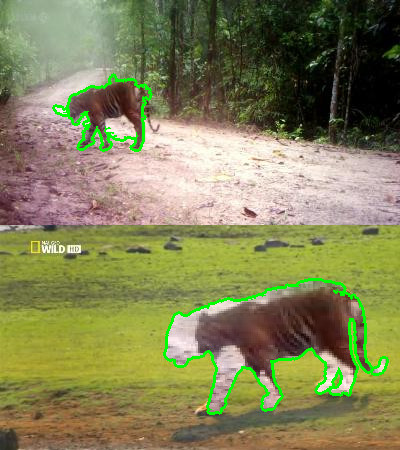}  &
\includegraphics[scale =0.25]{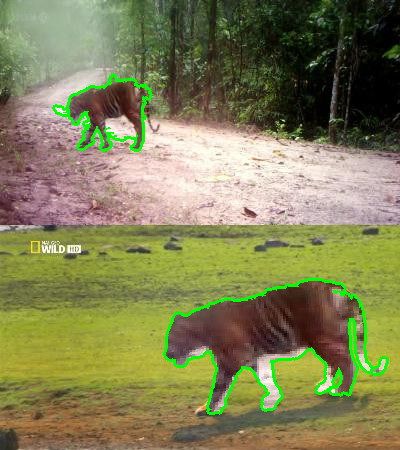}  &
\includegraphics[scale =0.25]{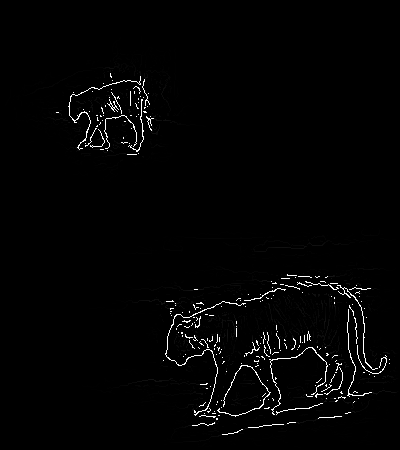}  \\
\small{(a)} & \small{(b)} & \small{(c)} & \small{(d)} & \small{(e)} \\
\end{tabular}
\end{center}
 \caption{\small{
Aligning sequences with similar foreground motion. We first estimate
a foreground mask (green) using motion segmentation (a). We then fit a 
homography to matches between point trajectories (b, sec.~\ref{sec:homogtraj}).
In (c) we project the foreground pixels in the first sequence 
(top) onto the second (bottom) with the recovered homography.
This global, coarse mapping is often not accurate (note the
misaligned legs and head).
We refine it by fitting Thin-Plate Splines (TPS) to
edge points extracted from the foreground (e, sec.~\ref{sec:tps}).
The TPS mapping is non-rigid and provides a more accurate alignment 
for complex articulated objects (d).
}
}
\label{fig:alignment} \end{figure*}

Most computer vision systems cannot take advantage of the abundance of Internet video content as training data. This is because current algorithms typically learn under strong supervision and annotating video content is expensive. Our goal is to remove the need for expensive manual annotations and instead reliably recover spatiotemporal correspondences between deformable objects under weak supervision. For instance, given a collection of animal documentary videos, can we automatically match pixels on a tiger in one video to those on a different tiger in another video (Figs.~\ref{fig:teaser} and \ref{fig:alignment})?

Recovering point-to-point spatiotemporal correspondences across videos is powerful
because it enables to assemble a collection of \emph{aligned}
foreground masks from a collection of videos of the same object class (Fig.~\ref{fig:teaser}).
Accomplishing this task in the presence of significant object appearance variations is
particularly important in order to capture the richness of the visual concept (\eg, different coloring and textures of an animal).
Achieving this could replace the expensive manual annotations required by several 
popular methods for learning visual concepts~\cite{dalal05cvpr,felzenszwalb10pami,
viola:nips05,
cinbis13iccv,wang13iccv,girshick14cvpr},
including methods that require annotations at the part level~\cite{Felzenszwalb03pictorialstructures,BourdevMalikICCV09,azizpour12eccv}.
Additionally, it can enable novel applications, such as replacing one instance of
an object with a suitable instance from a different video (like
the orange and the white tiger in Fig.~\ref{fig:teaser}).

Instances of the same class in different videos exhibit
large variations in appearance.
Hence, traditional methods for matching still images using local
appearance descriptors~\cite{barnes10eccv,liu08eccv,Hartley00,lowe04ijcv} 
typically do not find reliable correspondences.
We do this more effectively by aligning short temporal intervals where the objects exhibit consistent motion patterns. We exploit the characteristic motion of an object
class (\eg, a tiger's prowl) to identify suitable correspondences, 
and combine motion and edge features to align them with great accuracy.

We present a new technique to align two sequences of frames spatiotemporally using a set of Thin Plate Splines (TPS),
an expressive non-rigid mapping that has primarily been used 
for registration~\cite{Chui03} and shape matching~\cite{ferrari10ijcv}
in still images.
We extend these ideas to video by fitting a TPS that varies smoothly in time to minimize the distance between edge points in corresponding frames from the two sequences.

We evaluate our method on a new set of ground-truth annotations:
19 landmarks (\eg, left eye,
front left knee, neck, etc.) for two classes (horses and tigers).
We annotated $\sim$100 video shots per class, for a total of
$\sim$35,000 annotated frames (25 minutes of video).
The tiger shots come from a dataset of high-quality nature
documentaries filmed by professionals~\cite{delpero15cvpr}. The horse shots are sourced
from the YouTube-Objects dataset~\cite{prest12cvpr}, which are primarily
low-resolution footage filmed by amateurs.
This enables quantitative analysis on a large scale in two different settings.
Experiments show that our method recovers around a 1000 pairs of correctly aligned sequences
from 100 real-world video shots of each class.
As the recovered alignment is between {\em sequences}, this amounts to having correspondences between 10,000 pairs of frames.
This significantly outperforms the traditional approach of matching SIFT keypoints~\cite{lowe04ijcv} and the popular SIFT Flow algorithm~\cite{liu08eccv}.

The contributions of our work are:
(1) a weakly supervised system that goes from a large collection of unstructured video of an object class to a tight network of spatiotemporal correspondences between object instances;
(2) a method for aligning sequences of frames with consistent motion using TPS;
(3) publicly releasing the ground-truth annotations above. To our knowledge, this is the largest benchmark for sequence alignment to date.



\section{Related work}
\paragraph{Still image alignment.} Most works on spatial alignment focus on
matching between images for a variety of applications such
as multi-view reconstruction~\cite{seitz2006comparison},
image stitching~\cite{Brown2007}, and object instance
recognition~\cite{ferrari06ijcv,lowe04ijcv}.
The traditional approach identifies candidate matches using a local appearance descriptor (\eg,
SIFT~\cite{lowe04ijcv}) with global geometric verification performed using RANSAC~\cite{Fischler81,chum08pami}
or semi-local consistency checks~\cite{Schmid96,ferrari06ijcv,jegou08eccv}.
%
%
PatchMatch~\cite{barnes10eccv} and SIFT Flow~\cite{liu08eccv} generalize this notion to match patches between 
semantically similar scenes.

\paragraph{Sequence alignment.}
Our method differs from previous work on sequence
alignment~\cite{caspi00cvpr,caspi06ijcv,ukrainitz06eccv} in several ways.
First, we find correspondences between different scenes, rather than between
different views of the same scene~\cite{caspi00cvpr,caspi06ijcv}.
%
While the method in \cite{ukrainitz06eccv} is able to align actions across different scenes by directly maximizing local space-time correlations, it cannot handle the large intra-class appearance variations and diverse camera motions present in our videos.
As another key difference, all above approaches require temporally pre-segmented videos. Instead, we operate on
unsegmented videos and our method automatically finds which portions of each
can be spatiotemporally aligned.
Finally, these works have been evaluated only qualitatively
on 5-10 pairs of sequences, providing no quantitative analysis.

In the context of video action recognition,
there has been work on matching of spatiotemporal templates to
actor silhouettes~\cite{WeizmannActions,yilmaz05cvpr} or groupings
of supervoxels~\cite{ke07iccv}.
Our work is different because we map pixels from one unstructured video to another.
The method in~\cite{Jain2013} mines discriminative space-time patches and
matches them across videos.  It focuses on rough alignment using sparse
matches (typically one patch per clip), while we seek a finer, non-rigid spatial
alignment.
Other works on sequence alignment focus on temporal rather than spatial alignment~\cite{rao03iccv} or
target a very specific application, like aligning presentation slides to videos of the corresponding 
lecture~\cite{fan11tip}.

\begin{figure*}[t]
\begin{center}
\includegraphics[scale =0.085]{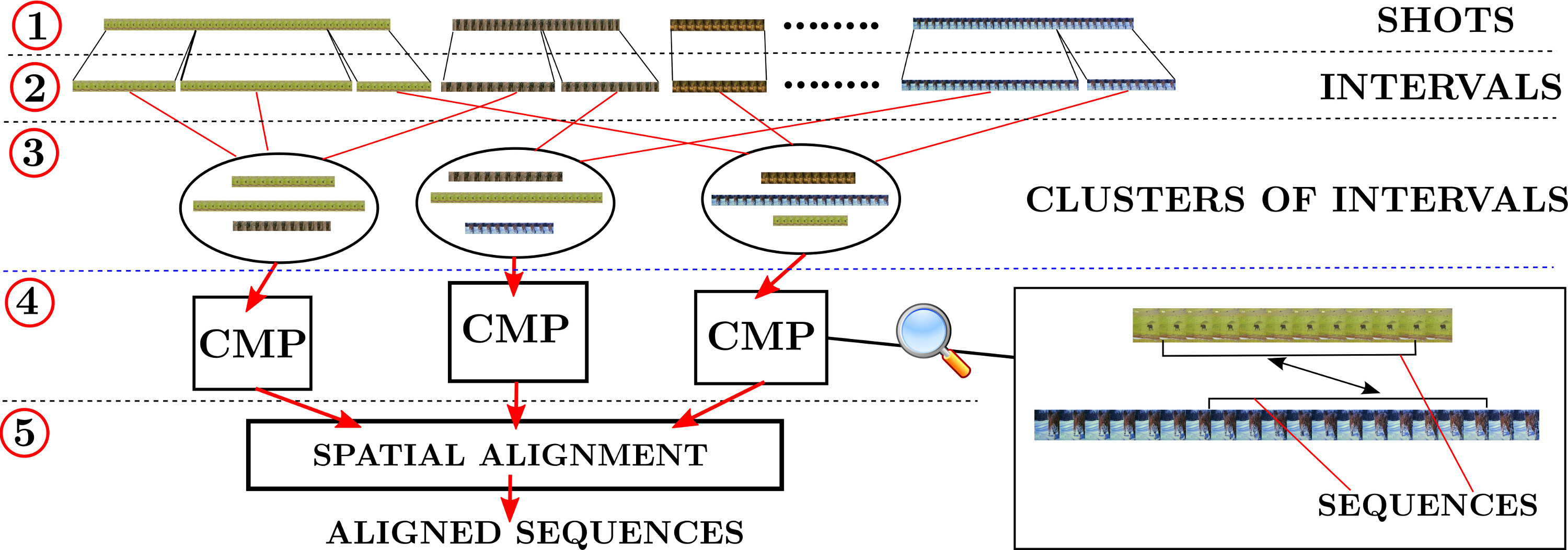}
\end{center}
 \caption{\small{
Overview of our method. The input is a collection of shots showing the
same class (1). Each shot (which can be of any length) is partitioned into shorter temporal
intervals of 10--200 frames (2), which are then clustered together (3) using
motion cues. (\cite{delpero15cvpr} shows that using intervals shorter than
the original shots finds more compact clusters.) The clusters effectively limit
the search space: we extract CMPs only from pairs of intervals
in the same cluster (4). For each pair, we extract CMPs from all
possible pairs of sequences of fixed length (10 frames). An example of a pair is shown
in the bottom right (sec.~\ref{sec:overview} and Fig.~\ref{fig:candidates}). 
Last (5), we align the two sequences of each CMP (see sec.~\ref{sec:homog} and~\ref{sec:tps}).
}}
\label{fig:overview}
\end{figure*}

\paragraph{TPS alignment.} 
TPS were developed as a general purpose smooth functional mapping
for supervised learning~\cite{wahba90tps}.
TPS have been used for non-rigid point matching between still images~\cite{Chui03},
and to match shape models to images~\cite{ferrari10ijcv}. 
The computer graphics community recently proposed semi-automated
video morphing using TPS~\cite{liao14cgf}.
However, this method requires manual point correspondences as input, 
and it matches image brightness directly.

\paragraph{Learning from videos.}
A few recent works exploit video as a source of training data 
for object class detectors~\cite{prest12cvpr,Tang2013}. 
However, their use of video is limited to segmenting objects from their background. 
Ramanan \etal~\cite{ramanan06pami} build a simple 2D pictorial structure model of an 
animal from one video. None of these methods find spatiotemporal correspondences between 
different instances of a class.

\section{System architecture}
\label{sec:overview}
Our method takes as input a large set of video shots containing instances of an object class (\eg tigers).  
These input shots are neither temporally segmented nor pre-aligned in any way.
The output is a collection of pairwise correspondences between frames.
Each correspondence is both temporal, \ie we find
correspondences between frames in different shots,
and spatial, \ie we recover the transformation 
mapping points between the two frames (Fig.~\ref{fig:teaser}, bottom).

\paragraph{Overview.} 
Fig.~\ref{fig:overview} shows an overview of our system.  The key
idea is to first identify pairs of frame sequences from two videos
that exhibit consistent foreground motion.  
For this, we use~\cite{papazoglou13iccv} to
extract foreground masks from each shot using motion cues and~\cite{delpero15cvpr}
to cluster short intervals with similar foreground motion.
Within each cluster, we identify pairs of sequences of fixed length
$T=10$ containing similar foreground motion;
we term these \emph{consistent motion pairs} (CMPs).
By focusing on similar motion, CMPs provide reliable correspondences
even across object instances with very different appearance
(such as the white and orange tigers in Fig.~\ref{fig:teaser} or the
cub and adult in Fig.~\ref{fig:bboxvsransac}).
These are fed to the next stage, which
spatiotemporally aligns the two sequences in each CMP.

\vspace{-6pt}
\paragraph{Foreground masks.}
We use the fast video segmentation technique~\cite{papazoglou13iccv} to automatically segment the foregound object from the background. These foreground masks remove confusing features on the background and facilitate the alignment process.

\begin{figure}
\begin{center}
\includegraphics[scale =0.3]{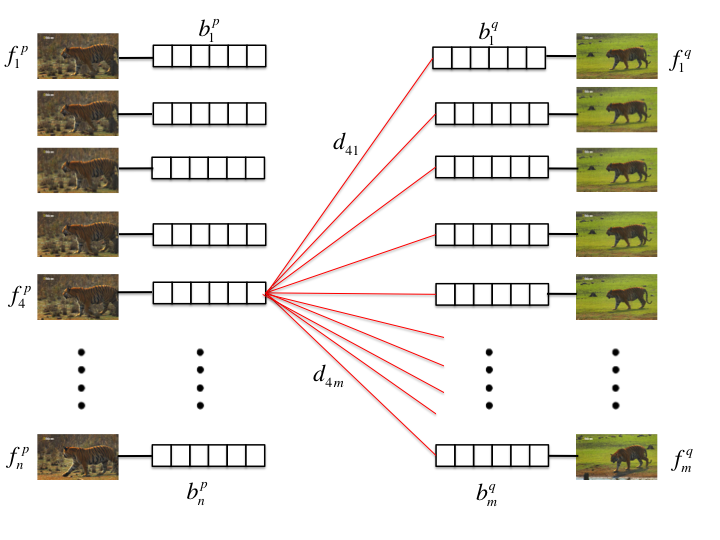}
\end{center}
 \caption{\small{
Extracting CMPs from two intervals.
First, we approximate the pairwise distance between frames
as the histogram distance between their BoWs (which contains all
motion descriptors through the frame, sec.~\ref{sec:overview}). Then we keep as CMPs
the top scoring pairs of sequences of length $T$ with respect to
(\ref{eq:candidates}). For the intervals above,
the number of pairs of sequences to examine is $(n-T)\cdot (m-T)$.
}}
\label{fig:candidates}
\end{figure}

\vspace{-6pt}
\paragraph{CMP extraction.}
Attempting to spatially align all possible pairs of sequences
would be prohibitively expensive (there are over a billion in
just 20 minutes of video). Clustering based on motion with~\cite{delpero15cvpr} 
significantly limits the search space. We prune further by considering only the top 
10 ranked pairs between two intervals $p$ and $q$ in the same cluster according to
the following metric (Fig.~\ref{fig:candidates}).
We describe each frame using a bag of words (BoW) over the
Trajectory Shape and Motion Boundary Histogram
descriptors~\cite{wang_ICCV_2013} of trajectories starting in that frame. Let
$d_{ij}$ be the histogram intersection between the BoWs for frame $i$ in $p$
and frame $j$ in $q$. The similarity between the $T$-frame sequence pair
starting at $i$ and $j$ is
%
\begin{equation}
\mathrm{s}\!\left([f_{i}^{p},\ldots,f_{i+T-1}^{p}]  ,  [f_{j}^{q},\ldots,f_{j+T-1}^{q}]\right)=\sum_{t=0}^{T-1}d_{(i+t)(j+t)} \, .
\label{eq:candidates}
\end{equation}
This measure preserves the temporal order of the frames, whereas a BoW
aggregated over the whole sequences would not.  
We found this scheme extracts CMPs that reliably show similar
foreground motion and form good candidates for spatial alignment.


\paragraph{Sequence alignment.}
We have explored a variety of approaches for sequence alignment and report on two representative methods here.
The first is a coarse rigid alignment generated by fitting a single homography
to foreground trajectory descriptors matched across the two sequences
(Sec.~\ref{sec:homog}).
The second approach fits a non-rigid TPS mapping to edge points extracted from the foreground regions of each frame.  This TPS is allowed to deform smoothly through time through the sequence (Sec.~\ref{sec:tps}).
Our experiments confirm that the more flexible model outperforms the rigid alignment (Sec.~\ref{sec:evaluation}).


\section{Rigid sequence alignment}
\label{sec:homog}  
Traditionally, homographies are used to model the mapping between two 
still images, and are estimated from a set
of noisy 2D point correspondences~\cite{Hartley00}.
We consider instead the problem of estimating a homography from
trajectories correspondences between two sequences (in a CMP).
Below we first review the standard approach for still images, and then present 
our extensions. 

\subsection{Homography between still images}
\label{sec:homogstill}

A 2D homography $H_{uv}$ is a $3\times3$ matrix that
can be determined from four or 
more point correspondences $X_{u} \leftrightarrow X_{v}$ by solving
\begin{equation} X_u = H_{uv}X_v
\label{eq:fittinghomo}
\end{equation}
%
RANSAC~\cite{Fischler81} estimates a homography from a set
of putative correspondences $\mathcal{P}_{uv} = \{(x_u, y_u) \leftrightarrow
(x_v, y_v)\}$ that may include outliers. Traditionally, $\mathcal{P}_{uv}$
contains matches between local appearance descriptors, like SIFT~\cite{lowe04ijcv}.
At each iteration, a hypothesis is
generated by fitting a homography to four samples from
$\mathcal{P}_{uv}$; the computed homography with the smallest number of outliers
is kept.

\subsection{Homography between video sequences}
\label{sec:homogtraj}

\begin{figure}
\begin{center}
\includegraphics[scale = 0.78]{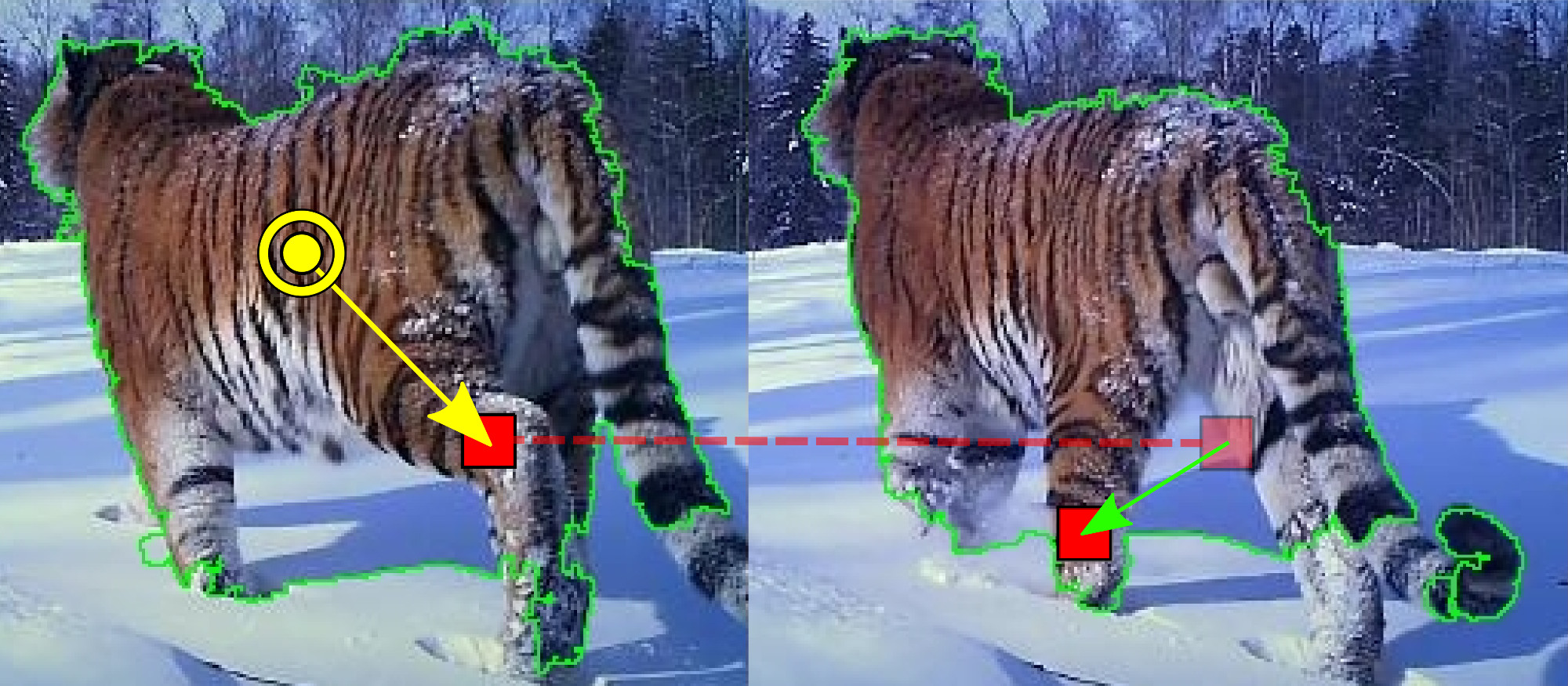}
\end{center}
 \caption{\small{
Modifying the TS descriptor. The TS descriptor is the concatenation
of the 2D displacement vectors (green) of a trajectory across consecutive frames.
This descriptor works well when aggregated in unordered representations
like Bag-of-Words~\cite{wang_ICCV_2013}, but matches found between
individual trajectories are not very robust. 
For example, the TS descriptors for the trajectories on the torso of a tiger walking are almost identical.
We make TS more discriminative by appending the vector (yellow) between the trajectory
and the center of mass of the foreground mask (green) in the frame where
the trajectory starts. We normalize this vector by the diagonal of the bounding box of the foreground mask
to preserve scale invariance.
 }}
\vspace{-6pt}
\label{fig:modifiedtraj}
\end{figure}

In video sequences, we use point trajectories as units for matching, instead of SIFT keypoints.
We extract trajectories in each sequence and match them using a modified Trajectory Shape (TS) descriptor~\cite{wang_ICCV_2013} (Fig.~\ref{fig:modifiedtraj}). 
We match each trajectory in the first sequence to its nearest neighbor 
in the second with respect to Euclidean distance. 
We use trajectories that are 10 frames long and only match those that start in the same frame in both sequences.
Each trajectory match provides 10 point correspondences (one per frame).

We consider two alternative ways to fit a homography to trajectory matches.
In the first, we treat the point correspondences generated by a single trajectory match independently during RANSAC. We call this strategy `Independent Matching' (IM).
In the second alternative, we sample four {\em trajectory} matches at each RANSAC iteration 
instead of four point correspondences.
We solve~\eqref{eq:fittinghomo} using the $4T$ associated point  
correspondences, in the least squares sense.
A trajectory match is considered an outlier 
only if fewer than half of its point correspondences are outliers.
We call this strategy `Temporal Matching' (TM).
TM encourages geometric consistency over the duration of the CMP.
Instead, IM could overfit to point correspondences from just a few
frames. Our experiments show that TM is superior to IM. 

\subsection{Using the foreground mask as a regularizer}
\label{sec:homogreg}
\begin{figure}
\begin{center}
\includegraphics[scale = 0.3]{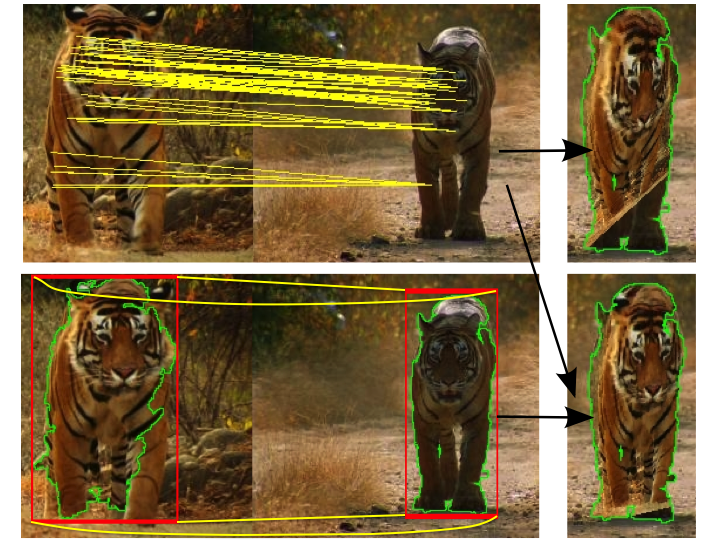}
\end{center}
 \caption{\small{
Top: Trajectory matches (yellow) often cover 
only part of the object (head and right leg here). 
Here, the homography overfit to correspondences
on the head, providing an incorrect mapping for the legs (right).
Bottom: Adding correspondences from the foreground
bounding boxes provides a more stable mapping (right).
The correspondences in the bottom
row are also found automatically by our method (no manual intervention needed).
 }}
\label{fig:bboxvsransac}
\end{figure}

\begin{figure}
\begin{center}
\includegraphics[scale = 0.21]{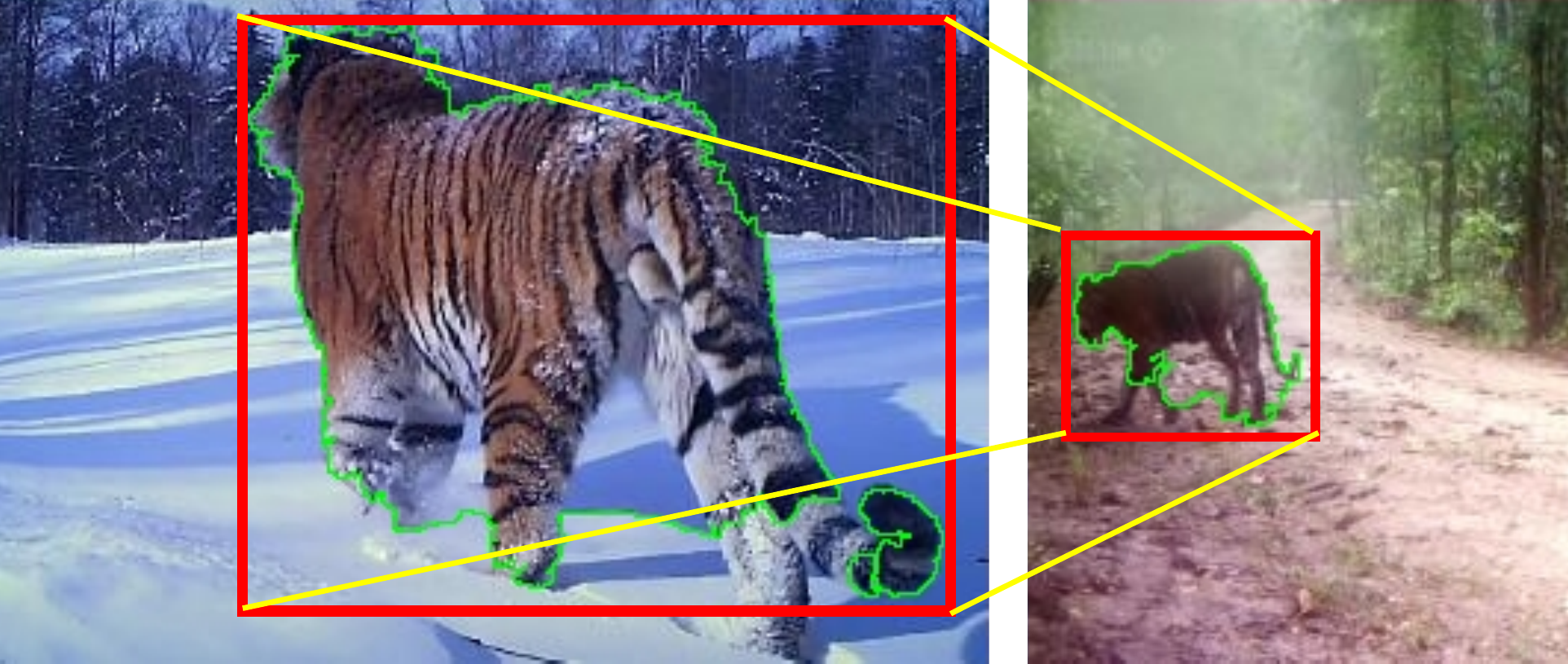}
\end{center}
 \caption{\small{
     Matching corners between the bounding boxes of the foreground mask provide
     additional point correspondences between the two sequences. While these
     correspondences are too coarse to provide a detailed spatial alignment
     between the sequences, and are sensitive to errors in the foreground
     segmentation (see Fig.~\ref{fig:tmvsfg}), they are useful as a regularizer
     when combined with other point correspondences.
 }}
\label{fig:bbox}
\end{figure}

The homography estimated from trajectories
tends to be inaccurate when the input
matches do not cover the entire foreground (Fig.~\ref{fig:bboxvsransac}).
To address this issue, we note that the bounding boxes of the foreground masks~\cite{papazoglou13iccv} provide
a coarse, global mapping (Fig.~\ref{fig:bbox}).
Specifically, we consider the correspondences between the bounding box
corners, which we call `foreground matches' ($F_{u}$, $F_{v}$).
These are included in Eq.~\eqref{eq:fittinghomo} as additional
point correspondences (four per frame):
\begin{equation}
\min \|H_{uv}X_{v}- X_{u}\| + \|H_{uv}F_{v}- F_{u}\|~~~.
\label{eq:fittinghomomodified}
\end{equation}
This form of regularization makes our method much more stable (Fig.~\ref{fig:bboxvsransac}).

\vspace{-10pt}
\section{Temporal TPS for sequence alignment}
\vspace{-4pt}
\label{sec:tps}

In this section we present a second approach to sequence alignment, based on time-varying thin plate splines (TTPS). Unlike the approach presented in the previous section, TTPS is a non-rigid mapping, which is more suitable for putting different object instances in correspondence.
We build on the popular TPS Robust Point Matching algorithm~\cite{Chui03}, originally developed to align sets of points between two still images (Sec.~\ref{sec:tpsrpm}).
We extend TPS-RPM to align two sequences of frames with a TPS that evolves smoothly over time (Sec.~\ref{sec:tpssequence}).


\subsection{TPS-RPM}
\label{sec:tpsrpm}

A TPS is a smooth, non-rigid mapping, $\mathrm{f}$, comprising an affine transformation $d$ and a non-rigid warp $w$. The mapping is a single closed-form function for the entire space,
with a smoothness term $\mathrm{L}(\mathrm{f})$ defined as the sum of the squares of the second
derivatives of $\mathrm{f}$ over the space~\cite{Chui03}.
Given two sets of points $\mathcal{U} = \{u_i\}$ and $\mathcal{V} = \{v_i\}$ in correspondence,
$\mathrm{f}$ can be estimated by minimizing
\begin{equation}
\label{eq:basictps}
\mathrm{E}(\mathrm{f}) = \sum_{i} || u_{i} - \mathrm{f}(v_{i})||^{2} + \lambda||\mathrm{L}(\mathrm{f})||.
\end{equation}
$\mathcal{U}$ and $\mathcal{V}$ are typically the position of detected image features (we use edge points, sec.~\ref{sec:tpssequence}).

As the point correspondences are typically not known beforehand, TPS-RPM jointly estimates $\mathrm{f}$
and a soft-assign correspondence matrix $M=\{m_{ij}\}$ 
by minimizing
\begin{equation}
\label{eq:tps-rpm}
\mathrm{E}(M, \mathrm{f}) = \sum_{i} \sum_{j} m_{ij}|| u_{i} - \mathrm{f}(v_{j})||^{2} + \lambda||\mathrm{L}(\mathrm{f})||.
\end{equation}
TPS-RPM alternates between updating $\mathrm{f}$ by keeping
$M$ fixed, and the converse. $M$ is continuous-valued, allowing
the algorithm to evolve through a continuous correspondence
space, rather than jumping around in the space
of binary matrices (hard correspondence).
It is updated by setting
$m_{ij}$ as a function of the distance between $u_{i}$
and $\mathrm{f}(v_{j})$~\cite{Chui03}. 
The TPS is updated by fitting $\mathrm{f}$ between
$\mathcal{V}$ and the current estimates $\mathcal{Y}$ of the corresponding points,
computed from $\mathcal{U}$ and $M$.
%
%
TPS-RPM and optimizes \eqref{eq:tps-rpm} in a deterministic annealing framework,
which allows TPS-RPM to find a good solution even when starting from a relatively poor initialization. \vitto{need to mention 'robust to outliers', or 'decides to match not all the points'}

\subsection{Temporal TPS}
\label{sec:tpssequence}

\begin{figure}
\begin{center}
\includegraphics[scale =0.32]{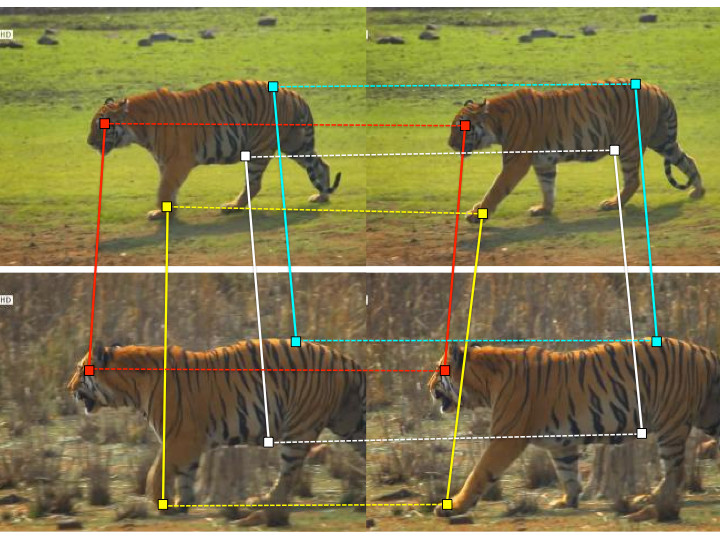} 
\end{center}
 \caption{\small{
Propagation using optical flow. In each sequence, we propagate
edge points extracted at time $t$ using optical flow,
independently in each sequence (dashed lines).
Our TTPS model (Sec.~\ref{sec:tpssequence}) enforces
that the correspondences between edge points at time $t$ (solid lines)
are consistent with their propagated version at time $t+1$.
}
}
\label{fig:propagation} \end{figure}

Our goal is to find a series of mappings $\mathcal{F}=\{\mathrm{f}^{1},\ldots,\mathrm{f}^{T} \}$,
one at each frame in the input sequences. We enforce temporal smoothness by
constraining each mapping to use a set of point correspondences that is consistent over time. 
Let $\mathcal{U}^t = \{u_i^t\}$ be a set of points for 
frame $t$ in the first sequence (with $\mathcal{V}^t$ defined analogously). This set
contains both edge points extracted in
$t$ as well as edge points extracted in other frames of the sequence and
propagated to $t$ via optical flow (Fig.~\ref{fig:propagation}). Each
$\mathcal{U}^t$ stores points in the same order such that $u_i^t$ and
$u_i^{\tau}$ are related by flow propagation.
We solve for $\mathcal{F}$ by minimizing
\begin{equation}
\label{eq:tpsflow}
\mathrm{E}(\mathcal{M}, \mathcal{F}) = \sum_{t}\! \left(\sum_{i}\! \sum_{j} m_{ij}^t|| u_{i}^{t} - \mathrm{f}^{t}(v_{j}^{t})||^{2} + \lambda||\mathrm{L}(\mathrm{f}^{t})|| \!\right) .
\end{equation}
subject to the constraint that $m_{ij}^t = m_{ij}^{\tau}$. That is, if two points
are in correspondence in frame $t$, they must still be in correspondence after being
propagated to frame $\tau$.

\paragraph{Inference.}
Minimizing (\ref{eq:tpsflow}) is very challenging. In practice, we find an
approximate solution by first using TPS-RPM to fit a separate TPS $\mathrm{f}^t$ to
the edge points extracted at time $t$ only. This is initialized with the homography
found in Sec.~\ref{sec:homogreg}.
$\mathrm{f}^{t}$ fixes the correspondences, which we use to estimate $\mathrm{f}^{\tau}$ in all other 
frames. 
We repeat this process starting in each frame, generating a total of $T$ TTPS candidates and keep the highest scoring one according to (\ref{eq:tpsflow}).
Thanks to this efficient approximate inference, we can apply TTPS to align
thousands of CMPs.

\begin{figure}[t]
\begin{center}
\setlength{\tabcolsep}{0.8pt}
\begin{tabular}{ccccc}
\footnotesize{\textbf{fg mask}} & \footnotesize{\textbf{all edges}} & \footnotesize{\textbf{fg edges}} & \footnotesize{\textbf{edges*DT}} \\
\includegraphics[scale =0.3]{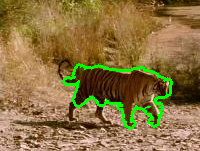}  &
\includegraphics[scale =0.3]{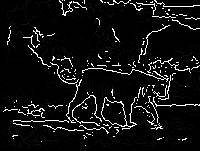}  &
\includegraphics[scale =0.3]{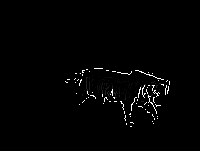}  &
\includegraphics[scale =0.3]{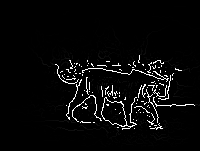}  \\
\small{(a)} & \small{(b)} & \small{(c)} & \small{(d)} \\
\end{tabular}
\end{center}
 \caption{\small{
Edge extraction. Using edges extracted from the entire image
confuses the TPS fitting due to background edge points (b).
Using only edges on the foreground mask (c) loses 
useful edge points if the mask is inaccurate, \eg the missing legs in (a).
We instead weigh the edge strength (b) by the Distance
Transform (DT) with respect to the foreground mask. This is robust to errors
in the mask, while pruning most background edges (d).
}
}
\label{fig:edges} \end{figure}

\paragraph{Foreground edge points.}
We extract edges using~\cite{dollar13iccv}.
We remove clutter edges far from the object by multiplying the edge strength of each point with the Distance Transform (DT) of the image with respect to the foreground mask (\ie, the distance of each pixel to the closest point on the mask). We prune points scoring $\leq 0.2$.
This removes most background edges, and is robust to cases where the mask does not cover the complete object (Fig~\ref{fig:edges}).
To accelerate the TTPS fitting process, after pruning we subsample the edge points to at most 1,000 per image.

\section{Evaluation}
\label{sec:evaluation}
We evaluate our method on $108$ shots of tigers
from a dataset of documentary nature footage~\cite{delpero15cvpr} 
and $98$ shots of horses from YouTube-Objects~\cite{prest12cvpr}, 
for a total of 17,000 frames per class (roughly 25 minutes of video).

\subsection{Evaluation protocol}
\label{sec:protocol}
\begin{figure}
\begin{center}
\includegraphics[scale = 0.24]{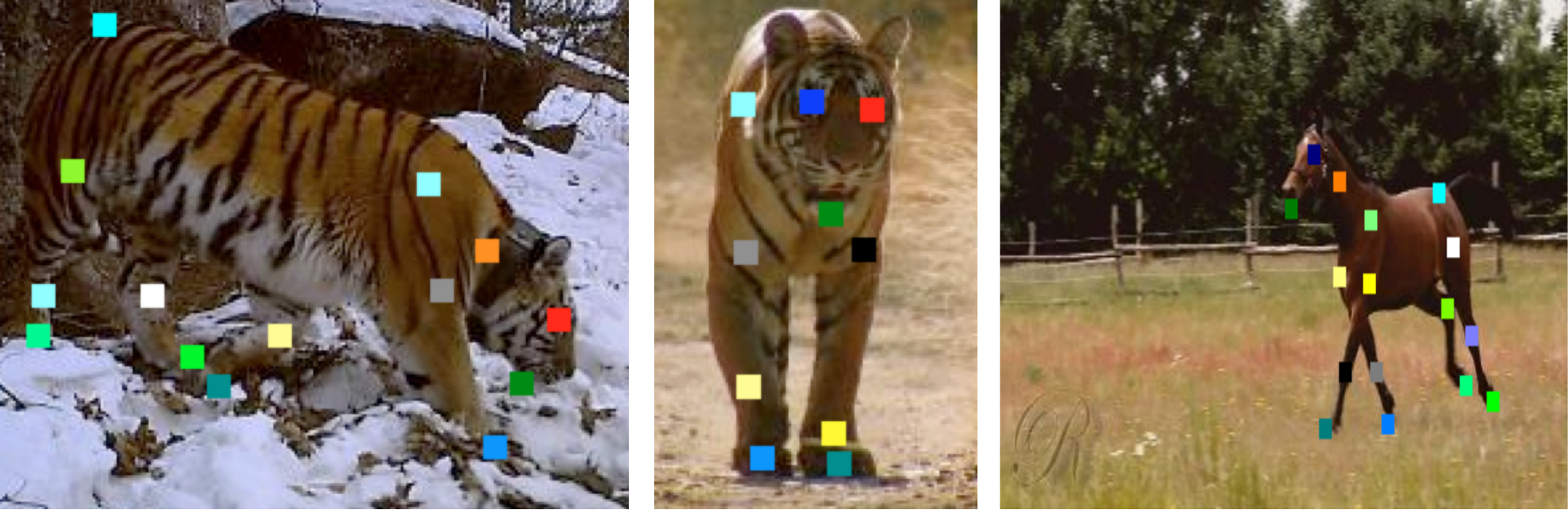}
\end{center}
 \caption{\small{
 Examples of annotated landmarks. A total of 19 points are marked when
 visible in over 17,000 frames for two different classes (horses and
 tigers). Our evaluation measure uses to landmarks to evaluate
 the quality of a sequence alignments (sec.~\ref{sec:protocol}).  }}
\label{fig:landmarks}
\end{figure}

\paragraph{Landmark annotations.}
In each frame, we annotate the 2D location
of 19 landmarks on each tiger/horse\footnote{If multiple are visible, we annotate the
animal closest to the camera.} (such as eyes, knees, chin, Fig.~\ref{fig:landmarks}).
We do not annotate occluded landmarks. We will make these annotations publicly available.
Unlike coarser annotations, such as bounding boxes,
landmarks enable evaluating the alignment of objects with
non-rigid parts with greater accuracy.

\paragraph{Evaluation measure.}
We evaluate the mapping found between the two sequences in a CMP as follows.
For each frame, we map each landmark in the first sequence onto the second and compute the Euclidean distance
to its ground-truth location.
The evaluation measure is the average between this distance and the reverse (\ie, the distance for landmarks mapped from the second sequence into the first).
We normalize the error by the scale of the object,
defined as the maximum distance between any two landmarks in the frame.
The overall error for a pair of sequences is the average error of all visible landmarks over all frames.

\begin{figure}
\begin{center}
\includegraphics[scale = 0.19]{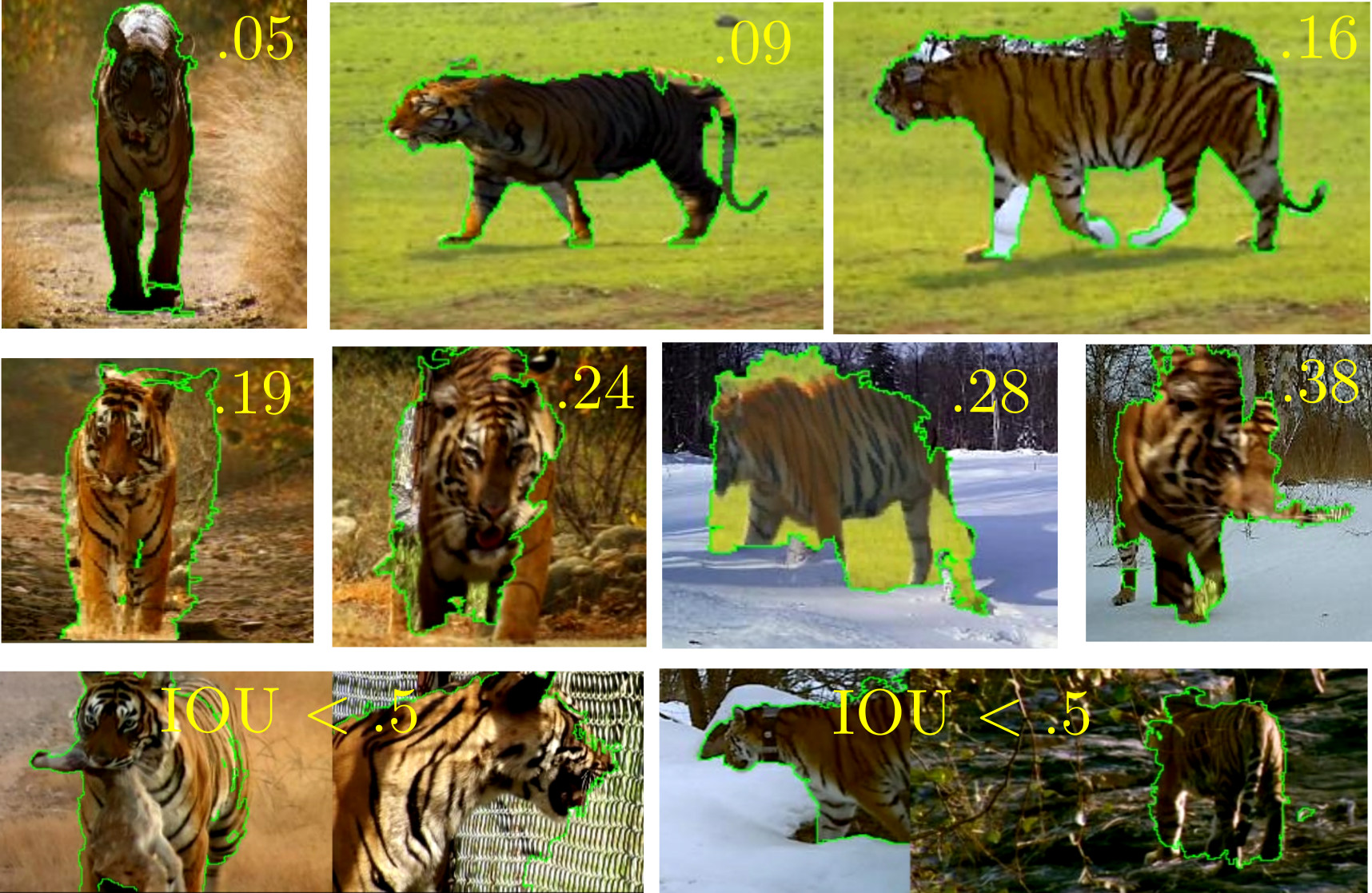}
\end{center}
 \caption{\small{
   Evaluation measure. We use the ground-truth landmarks
   to measure the alignment error of the mappings estimated by our method (sec.~\ref{sec:protocol}).
   As the error increases, the quality of the alignment clearly degrades.  Around $0.18$ the alignments contain
   some slight mistakes (\eg, the slightly misaligned legs in the top right image),
   but are typically acceptable. We consider a mapping incorrect also when the IOU of the 
   visible landmarks in the aligned pair is below $0.5$ (bottom row).
   }}
\label{fig:reperror}
\end{figure}

\ricco{Previous version in comments.}
After visual inspection of many sampled alignments (Fig.~\ref{fig:reperror}), we found that $0.18$ was a reasonable threshold for separating acceptable alignments from those with noticeable errors.
We count an alignment as \emph{correct} if the error is below this threshold and if the Intersection over
Union (IOU) of the two sets of visible landmarks in the sequence is above $0.5$ (to avoid rewarding accidental alignments of a few landmarks, bottom row of fig.~\ref{fig:reperror}).


\vspace{-4pt}
\subsection{Evaluating CMP extraction}
\vspace{-4pt}
\label{sec:evalcandidates}

First, we evaluate our method for CMP extraction 
in isolation (sec.~\ref{sec:overview}). Given a CMP, we use the ground-truth
landmarks to fit a homography, and check if it is correct according to the evaluation measure above.
If so, it means that it is in principle possible to align it (we call it \emph{alignable}).
Our method returns roughly 3000 CMP on the tiger data, of which $51\%$ are alignable.
As a baseline, we consider extracting CMPs by uniformly sampling sequences from pairs of shots. In this case, the percentage of alignable CMPs drops to $19\%$.
Results are similar on the the horse dataset: our method delivers $49\%$ alignable CMPs, vs $26\%$ by the baseline.

\subsection{Evaluating spatial alignment}
\label{sec:evalalignment}
We now evaluate various methods for automatic sequence alignment.
For each method, we generate a precision-recall curve as follows.
Let $n$ be the total number of CMPs returned by the method; $c$ the number of correctly aligned CMPs; and $a$ the total number of alignable CMPs (sec.~\ref{sec:evalcandidates}).
Recall is $c/a$, and precision is $c/n$.
Different operating points on the precision-recall curve are obtained by varying the maximum percentage of outliers allowed when fitting a homography.

\begin{figure}
\begin{center}
\includegraphics[scale = 0.22]{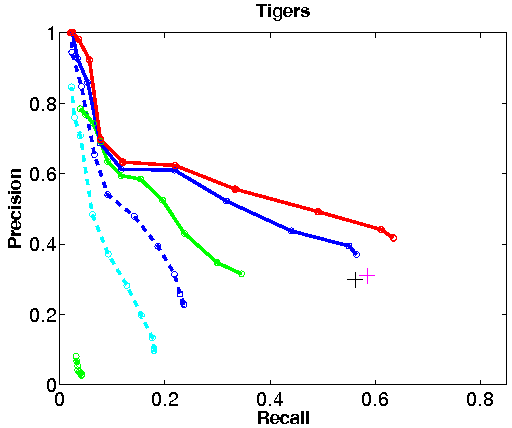}
\includegraphics[scale = 0.22]{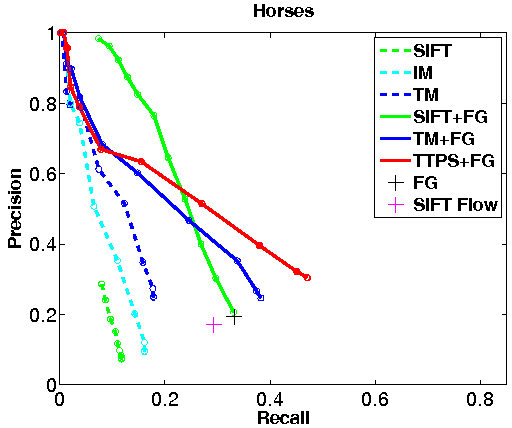}
\end{center}
 \caption{\small{
Evaluation of sequence alignment. We separately evaluate our method
on two classes, horses and tigers.
With no regularization, trajectory methods are superior to SIFT on both classes,
with TM performing better than IM. 
Adding regularization using the foreground matches
improves the performance of both TM and SIFT (compare
the dashed to the solid curves). 
TTPS clearly outperform all trajectory methods,
as well as SIFT Flow and the FG baseline (see text). }}
\label{fig:results}
\end{figure}

\paragraph{Comparison to other methods.}
We compare our method against SIFT Flow~\cite{liu08eccv}.
We use ~\cite{liu08eccv} to align each pair of frames from the
two sequences independently.
We restrict the algorithm to match only the bounding boxes of the foreground masks,
after rescaling them to be the same size (without these two steps,
performances significantly drop).

Further, we also compare to fitting a homography to SIFT matches found in the two sequences.
We use only keypoints on the foreground mask, and preserve
temporal order by matching only keypoints in corresponding frames.
We tested this method alone (SIFT), and by adding spatial regularization using 
the foreground masks (SIFT + FG, sec.~\ref{sec:homogreg}).

Finally, we consider a simple baseline that fits a homography to the bounding box of the foreground masks alone (FG).

\begin{figure}
\begin{center}
\setlength{\tabcolsep}{0.8pt}
\begin{tabular}{c c c c}
\footnotesize{\textbf{Video 1}} & \footnotesize{\textbf{Video 2}} & \footnotesize{\textbf{Homography}} & \footnotesize{\textbf{TPS}} \\
\includegraphics[scale =0.15]{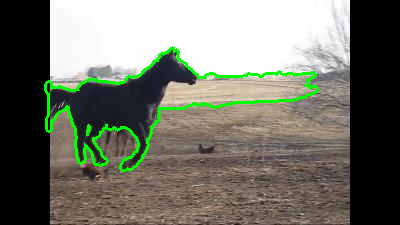}  &
\includegraphics[scale =0.15]{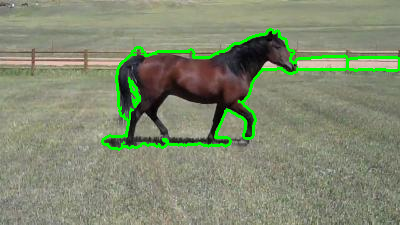}  &
\includegraphics[scale =0.15]{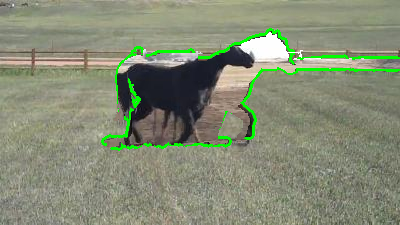}  &
\includegraphics[scale =0.15]{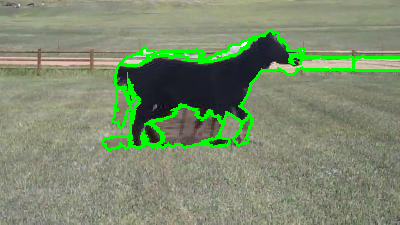} \\
\includegraphics[scale =0.15]{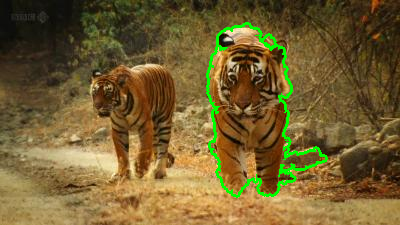}  &
\includegraphics[scale =0.15]{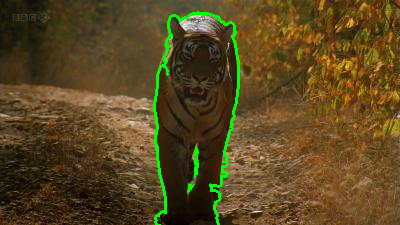}  &
\includegraphics[scale =0.15]{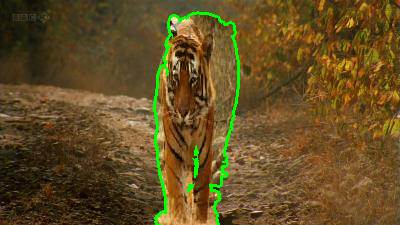}  &
\includegraphics[scale =0.15]{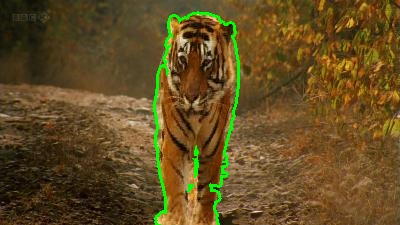} \\
\small{(a)} & \small{(b)} & \small{(c)} & \small{(d)} \\
\end{tabular}
\end{center}
 \caption{\small{
TTPS (d) provide a more accurate alignment for complex articulated objects than
homographies (c).}
}
\label{fig:tpsvshomo} \end{figure}

\begin{figure}
\begin{center}
\includegraphics[scale = 0.38]{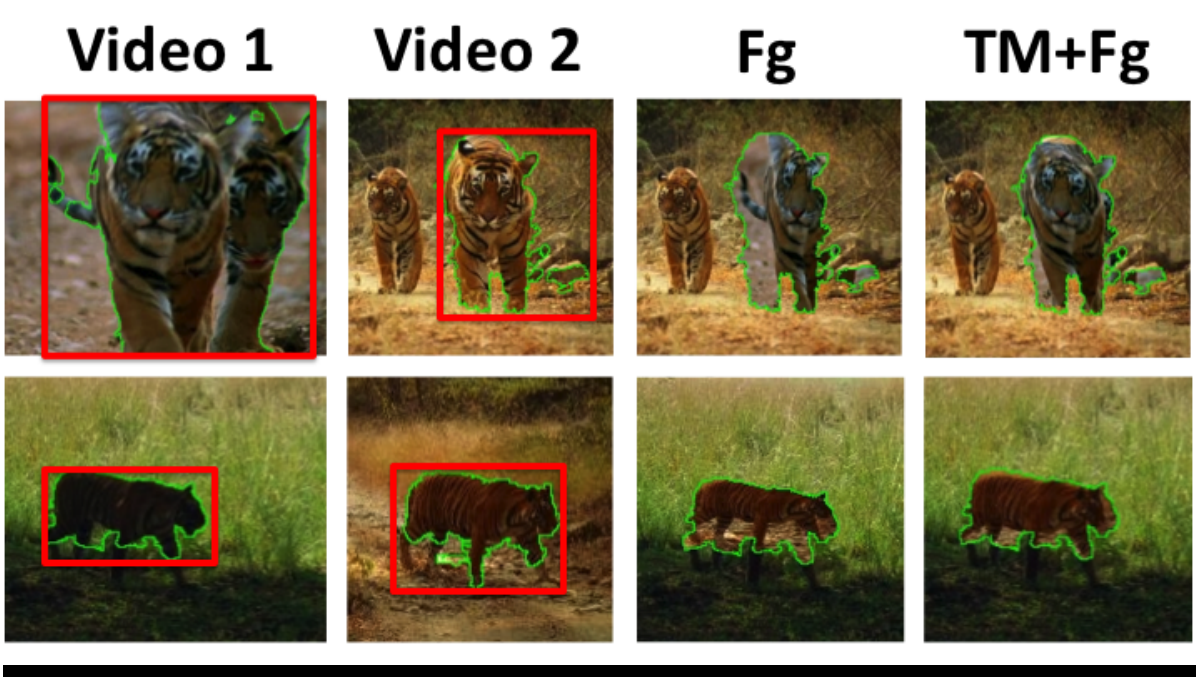}
\includegraphics[scale = 0.38]{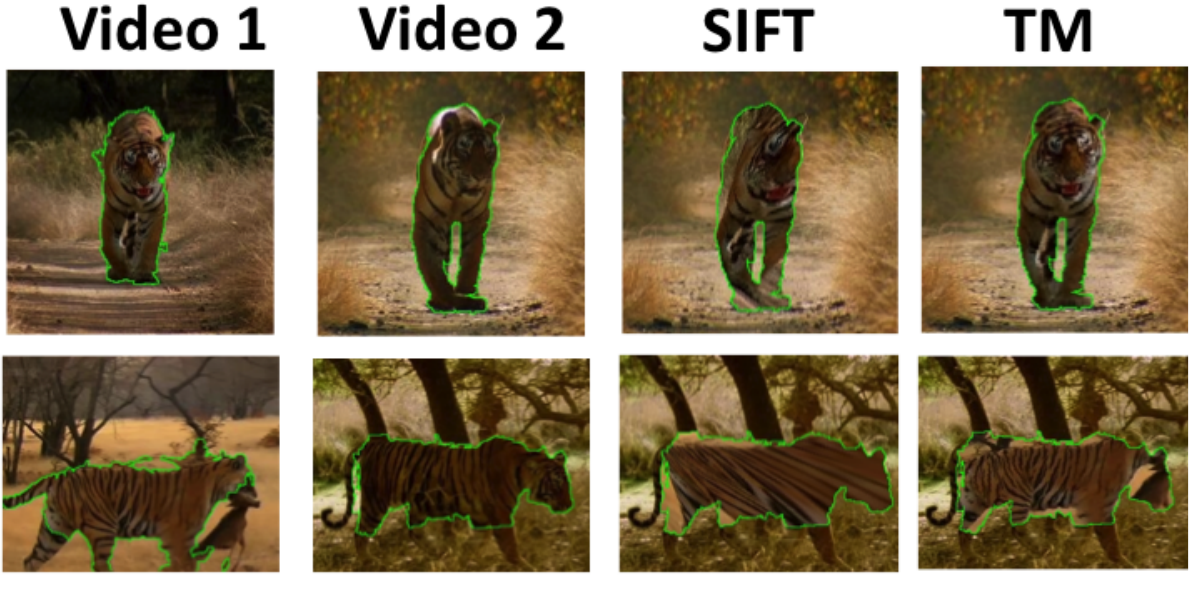}
\end{center}
 \caption{\small{
 Top two rows: Estimating the homography from the foreground masks alone fails
 when the bounding boxes are not tight around the objects (first-second columns). 
 Adding trajectories (TM+FG) is more accurate (fourth column, sec.~\ref{sec:homogtraj}).
 Bottom two rows: the striped texture of tigers often confuses estimating
 the homography from SIFT keypoint matches (third row). 
 On this class, using trajectories (TM)
 often performs better.
 }}
\label{fig:tmvsfg}
\end{figure}

\vspace{-6pt}
\paragraph{Analysis of rigid alignment.}
Both trajectory methods (TM, IM, sec.~\ref{sec:homogtraj}) are superior to SIFT on both classes, with TM performing better than IM (Fig.~\ref{fig:results}).
Adding spatial regularization with the foreground masks (+FG) improves the performance of both TM and SIFT.
SIFT performs poorly on tigers, since the striped texture confuses matching SIFT
keypoints (Fig.~\ref{fig:tmvsfg}, bottom).
Trajectory methods work somewhat better on tigers than horses due to the poorer quality of YouTube video (\eg low resolution, shaky camera, abrupt pans).
As a result of these factors, TM+FG clearly outperforms SIFT+FG on tigers, but it is somewhat worse on horses.
%

\paragraph{Analysis of TTPS.}
The time-varying TPS model (TTPS+FG, sec.~\ref{sec:tps}) significantly improves upon its initialization (TM+FG) on both classes. 
On tigers, it is the best method overall, as its precision-recall is above all other curves for the entire range.
On horses, the SIFT+FG and TTPS+FG curves intersect. However, TTPS+FG achieves a higher Average Precision (\ie the area under the curve): 0.265 vs 0.235.

The SIFT Flow software~\cite{liu08eccv} does not produce scores comparable across CMPs, so we cannot produce a full precision-recall curve. At the level of recall of SIFT Flow, TTPS achieves +0.2 higher precision on tigers, and +0.3 on horses. We also note that TM and TM+FG are closely 
related to the method for fitting homographies
to trajectories in~\cite{caspi06ijcv}. TM+FG augments~\cite{caspi06ijcv} in several ways 
(automatic CMP extraction, modified TS desriptor, regularization with the foreground masks), 
but is still inferior to TTPS+FG. \luca{I put caspi here instead
of at the end of the last paragraph.}
Last, TTPS also achieves a significantly higher precision than the FG baseline. 
This shows that our method is robust to errors in the foreground masks.
In supplemental material we provide example head-to-head qualitative results, showing that TTPS alignents 
typically look more accurate than the other methods (Fig.~\ref{fig:tpsvshomo}).

For the tiger class, out of all CPMs returned by TTPS (rightmost point on the curve),
$1,000$ of them are correctly aligned tiger (\ie $10,000$ frames). The precision at this point is $0.5$, i.e. half of the returned CMPs are correctly aligned.
For the horse class, TTPS returns 800 correctly aligned CMPs, with precision $0.35$.


\section{Discussion}
We present a method that automatically extracts dense spatiotemporal
correspondences from a collection of videos showing a particular object class.
Our pipeline consumes raw video, without the need for manual annotations or
temporal segmentation. Using motion as the primary signal for identifying
correspondences allows us to match sequences despite significant appearance
variation. Ultimately, the thin plate spline matching results in
temporally-stable, high-quality alignments for thousands of sequence pairs.



Our method is not limited to a particular class of object but instead applies
to any objects that exhibit consistency in behavior and thus exhibit the same
characteristic motion patterns across different observations. The
correspondences we find can be used to learn a
general model of the object class without requiring any human supervision beyond
video-level object class labels. Additionally, they can enable
novel applications, such as replacing an instance of an object 
with an instance from a different video, or retrieving videos in a 
collection that tightly match the motion of the object in a query video.

\paragraph{Acknowledgments.}
We are very grateful to Anestis Papazoglou for helping with the data collection,
and to Shumeet Baluja for his helpful comments. This work was partly funded by a Google Faculty Research Award,
and by ERC Starting Grant ``Visual Culture for Image Understanding''.


{\small
\bibliographystyle{ieee}
\bibliography{../../bibtex/shortstrings,../../bibtex/calvin,../../bibtex/vggroup}

\begin{thebibliography}{10}\itemsep=-1pt

\bibitem{azizpour12eccv}
H.~Azizpour and I.~Laptev.
\newblock Object detection using strongly-supervised deformable part models.
\newblock In {\em ECCV}, 2012.

\bibitem{barnes10eccv}
C.~Barnes, E.~Shechtman, D.~Goldman, and A.~Finkelstein.
\newblock The generalized patchmatch correspondence algorithm.
\newblock In {\em ECCV}, 2010.

\bibitem{BourdevMalikICCV09}
L.~Bourdev and J.~Malik.
\newblock Poselets: Body part detectors trained using 3d human pose
  annotations.
\newblock In {\em ICCV}, 2009.

\bibitem{Brown2007}
M.~Brown and D.~Lowe.
\newblock Automatic panoramic image stitching using invariant features.
\newblock {\em IJCV}, 74(1), 2007.

\bibitem{caspi00cvpr}
Y.~Caspi and M.~Irani.
\newblock A step towards sequence-to-sequence alignment.
\newblock In {\em CVPR}, 2000.

\bibitem{caspi06ijcv}
Y.~Caspi, D.~Simakov, and M.~Irani.
\newblock Feature-based sequence-to-sequence matching.
\newblock {\em IJCV}, 68(1):53--64, 2006.

\bibitem{Chui03}
H.~Chui and A.~Rangarajan.
\newblock A new point matching algorithm for non-rigid registration.
\newblock {\em CVIU}, 89(2-3):114--141, Feb 2003.

\bibitem{chum08pami}
O.~Chum and J.~Matas.
\newblock Optimal randomized ransac.
\newblock {\em IEEE Trans. on PAMI}, 2008.

\bibitem{cinbis13iccv}
R.~Cinbis, J.~Verbeek, and C.~Schmid.
\newblock Segmentation driven object detection with fisher vectors.
\newblock In {\em ICCV}, 2013.

\bibitem{dalal05cvpr}
N.~Dalal and B.~Triggs.
\newblock {H}istogram of {O}riented {G}radients for human detection.
\newblock In {\em CVPR}, 2005.

\bibitem{delpero15cvpr}
L.~Del~Pero, S.~Ricco, R.~Sukthankar, and V.~Ferrari.
\newblock Articulated motion discovery using pairs of trajectories.
\newblock In {\em CVPR}, 2015.

\bibitem{dollar13iccv}
P.~Dollar and C.~Zitnick.
\newblock Structured forests for fast edge detection.
\newblock In {\em ICCV}, 2013.

\bibitem{fan11tip}
Q.~Fan, K.~Barnard, A.~Amir, and A.~Efrat.
\newblock Robust spatio-temporal matching of electronic slides to presentation
  videos.
\newblock {\em {IEEE} Transactions on Image Processing}, 20(8):2315--2328,
  2011.

\bibitem{felzenszwalb10pami}
P.~Felzenszwalb, R.~Girshick, D.~McAllester, and D.~Ramanan.
\newblock Object detection with discriminatively trained part based models.
\newblock {\em IEEE Trans. on PAMI}, 32(9), 2010.

\bibitem{Felzenszwalb03pictorialstructures}
P.~F. Felzenszwalb and D.~P. Huttenlocher.
\newblock Pictorial structures for object recognition.
\newblock {\em IJCV}, 61(1):55--79, 2005.

\bibitem{ferrari10ijcv}
V.~Ferrari, F.~Jurie, and C.~Schmid.
\newblock From images to shape models for object detection.
\newblock {\em IJCV}, 87(3), 2010.

\bibitem{ferrari06ijcv}
V.~Ferrari, T.~Tuytelaars, and L.~Van~Gool.
\newblock Simultaneous object recognition and segmentation from single or
  multiple model views.
\newblock {\em IJCV}, 67(2):159--188, 2006.

\bibitem{Fischler81}
M.~A. Fischler and R.~C. Bolles.
\newblock Random sample consensus: {A} paradigm for model fitting with
  applications to image analysis and automated cartography.
\newblock {\em Comm. ACM}, 24(6):381--395, 1981.

\bibitem{girshick14cvpr}
R.~Girshick, J.~Donahue, T.~Darrell, and J.~Malik.
\newblock Rich feature hierarchies for accurate object detection and semantic
  segmentation.
\newblock In {\em CVPR}, 2014.

\bibitem{WeizmannActions}
L.~Gorelick, M.~Blank, E.~Shechtman, M.~Irani, and R.~Basri.
\newblock Actions as space-time shapes.
\newblock {\em IEEE Trans. on PAMI}, 29(12):2247--2253, December 2007.

\bibitem{Hartley00}
R.~I. Hartley and A.~Zisserman.
\newblock {\em Multiple View Geometry in Computer Vision}.
\newblock Cambridge University Press, 2000.

\bibitem{Jain2013}
A.~Jain, A.~Gupta, M.~Rodriguez, and L.~Davis.
\newblock Representing videos using mid-level discriminative patches.
\newblock In {\em CVPR}, 2013.

\bibitem{jegou08eccv}
H.~Jegou, M.~Douze, and C.~Schmid.
\newblock Hamming embedding and weak geometric consistency for large-scale
  image search.
\newblock In {\em ECCV}, 2008.

\bibitem{ke07iccv}
Y.~Ke, R.~Sukthankar, and M.~Hebert.
\newblock Event detection in crowded videos.
\newblock In {\em ICCV}, 2007.

\bibitem{liao14cgf}
J.~Liao, R.~S. Lima, D.~Nehab, H.~Hoppe, and P.~V. Sander.
\newblock Semi-automated video morphing.
\newblock In {\em Eurographics Symposium on Rendering}, 2014.

\bibitem{liu08eccv}
C.~Liu, J.~Yuen, A.~Torralba, J.~Sivic, and W.~Freeman.
\newblock {SIFT Flow}: Dense correspondence across different scenes.
\newblock In {\em ECCV}, 2008.

\bibitem{lowe04ijcv}
D.~Lowe.
\newblock Distinctive image features from scale-invariant keypoints.
\newblock {\em IJCV}, 60(2):91--110, 2004.

\bibitem{papazoglou13iccv}
A.~Papazoglou and V.~Ferrari.
\newblock Fast object segmentation in unconstrained video.
\newblock In {\em ICCV}, December 2013.

\bibitem{prest12cvpr}
A.~Prest, C.~Leistner, J.~Civera, C.~Schmid, and V.~Ferrari.
\newblock Learning object class detectors from weakly annotated video.
\newblock In {\em CVPR}, 2012.

\bibitem{ramanan06pami}
D.~Ramanan, A.~Forsyth, and K.~Barnard.
\newblock Building models of animals from video.
\newblock {\em IEEE Trans. on PAMI}, 28(8):1319 -- 1334, 2006.

\bibitem{rao03iccv}
C.~Rao, A.~Gritai, and M.~Shah.
\newblock View-invariant alignment and matching of video sequences.
\newblock In {\em ICCV}, 2003.

\bibitem{Schmid96}
C.~Schmid and R.~Mohr.
\newblock Combining greyvalue invariants with local constraints for object
  recognition.
\newblock Technical report, INRIA Rh\^one-Alpes, Grenoble, France, 1996.

\bibitem{seitz2006comparison}
S.~Seitz, B.~Curless, J.~Diebel, D.~Scharstein, and R.~Szeliski.
\newblock A comparison and evaluation of multi-view stereo reconstruction
  algorithms.
\newblock In {\em CVPR}, 2006.

\bibitem{Tang2013}
K.~Tang, R.~Sukthankar, J.~Yagnik, and L.~Fei-Fei.
\newblock Discriminative segment annotation in weakly labeled video.
\newblock In {\em CVPR}, 2013.

\bibitem{ukrainitz06eccv}
Y.~Ukrainitz and M.~Irani.
\newblock Aligning sequences and actions by maximizing space-time correlations.
\newblock In {\em ECCV}, 2006.

\bibitem{viola:nips05}
P.~A. Viola, J.~Platt, and C.~Zhang.
\newblock Multiple instance boosting for object detection.
\newblock In {\em NIPS}, 2005.

\bibitem{wahba90tps}
G.~Wahba.
\newblock {\em {Spline models for observational data}}, volume~59 of {\em
  CBMS-NSF Regional Conference Series in Applied Mathematics}.
\newblock SIAM, 1990.

\bibitem{wang_ICCV_2013}
H.~Wang and C.~Schmid.
\newblock Action recognition with improved trajectories.
\newblock In {\em ICCV}, 2013.

\bibitem{wang13iccv}
X.~Wang, M.~Yang, S.~Zhu, and Y.~Lin.
\newblock Regionlets for generic object detection.
\newblock In {\em ICCV}, pages 17--24. IEEE, 2013.

\bibitem{yilmaz05cvpr}
A.~Yilmaz and M.~Shah.
\newblock Actions as objects: A novel action representation.
\newblock In {\em CVPR}, 2005.

\end{thebibliography}
}

\end{document}